\documentclass{article}

\usepackage{arxiv}

\usepackage[utf8]{inputenc} 
\usepackage[T1]{fontenc}    
\usepackage{hyperref}       
\usepackage{url}            
\usepackage{booktabs}       
\usepackage{amsfonts}       
\usepackage{nicefrac}       
\usepackage{microtype}      
\usepackage{lipsum}
\usepackage{amsmath}
\usepackage{graphicx}
\usepackage{verbatim}
\graphicspath{ {./images/} }
\usepackage{todonotes}
\title{Copy this sentence.}

\newcommand{\etal}{\textit{et al.}}

\author{
  Vasileios Lioutas\\
  Department of Computer Science\\
  Carleton University\\
  Ottawa, ON \\
  \texttt{vasileios.lioutas@carleton.ca} \\
   \And
 Andriy Drozdyuk \\
  Department of Computer Science\\
  Carleton University\\
  Ottawa, ON \\
  \texttt{andriy.drozdyuk@carleton.ca} \\
}

\begin{document}
\maketitle

\begin{abstract}
Attention is an operation that selects some largest element from some set, where the notion of largest is defined elsewhere. Applying this operation to sequence to sequence mapping results in significant improvements to the task at hand. In this paper we provide the mathematical definition of attention and examine its application to sequence to sequence models. We highlight the exact correspondences between machine learning implementations of attention and our mathematical definition. We provide clear evidence of effectiveness of attention mechanisms evaluating models with varying degrees of attention on a very simple task: copying a sentence. We find that models that make greater use of attention perform much better on sequence to sequence mapping tasks, converge faster and are more stable.
\end{abstract}


\section{Introduction}
Sequence to sequence models are fundamental to many of the systems in machine learning and natural language processing in particular. Fundamentally, a sequence to sequence model is system that maps some sequence of length $k$ to another sequence of length $n$ where $k$ and $n$ are possibly different.

Textual data is inherently of variable lengths, e.g. sentences can have different number of words. However, neural networks can only handle fixed length inputs and outputs. Recurrent Neural Networks (RNNs) can accept variable length inputs. How do we use them for sequence to sequence mapping problem? In \cite{graves2013generating} authors demonstrate the use of recurrent neural networks for generating sequences, by taking the network's output and feeding it as the input. To use RNN for sequence to sequence mapping we employ a model with an encoder and a decoder \cite{sutskever2014sequence}. Encoder takes the first sequence to the encoded representation, while the decoder uses that representation to produce the target sequence.

In this work we compare how well different sequence to sequence approaches perform on a well defined task. To isolate the performance of different models we choose the simplest possible (to humans) sequence to sequence task: copying a sentence. Here the input sentence $S$ and the output $T$ should be equal, i.e.: $S=T$. While it may sound trivial, we find that many sequence to sequence models are unable to cope with this simple task adequately. The models we compare range from ``no-attention'', as originally presented in \emph{Sequence to Sequence Learning with Neural Networks} \cite{sutskever2014sequence}, to ``maximum attention''---from the  \emph{Attention is all you need} paper \cite{vaswani2017attention}. 

First use of attention in sequence to sequence mapping was proposed in \cite{bahdanau2014neural}, which uses LSTM hidden states and a simple feedforward neural network to learn the scoring function. Others \cite{gehring2017convolutional} have used CNN models to learn the scoring function. In \cite{parikh2016decomposable} and \cite{vaswani2017attention} the authors use attention without recurrent neural networks, allowing for faster learning, but necessitating manual position encoding to imbue the model with the knowledge about relative order of items in a sequence.

We start by defining the fundamental notion of \emph{attention}. In Section~\ref{sec:relatedwork} we describe some of the other related work that looked at sequence to sequence models. The  Section~\ref{sec:method} describes all of the models we used in detail. In Section~\ref{sec:experiment} we describe our dataset, evaluation metrics and the parameters we used for our implementations. We conclude with discussion and future work.

\subsection{Attention}
Here we introduce the most general definition of attention.

\emph{Attention} is an operation that selects some largest element from some set $X$ where the notion of largest is represented by some set $S$ of scores\footnote{Our definition of attention is due to lecture slides from Yongyi Mao at University of Ottawa.}:

\begin{equation}
    \hat{X} = X \cdot \textit{softmax}(S)
\end{equation}

Here the set $S = \{s(y_i) | y_i \in Y \}$ where $s: Y\to R$ is a score function that assigns a score to each $y_i \in Y$. Each $y_i$ is some \emph{evidence} according to which a particular $x_i$ is to be selected.
Since $s$ is a function, we can learn it (e.g. represent it as a neural network with some parameter $\theta$): 

\begin{equation}
S = s(Y^T; \theta)
\label{eq:attention}
\end{equation}

In the degenerate case when $X = Y$ the operation is called \emph{self-attention}.

\begin{figure}[htbp]
\centering
\includegraphics[width=0.7\textwidth]{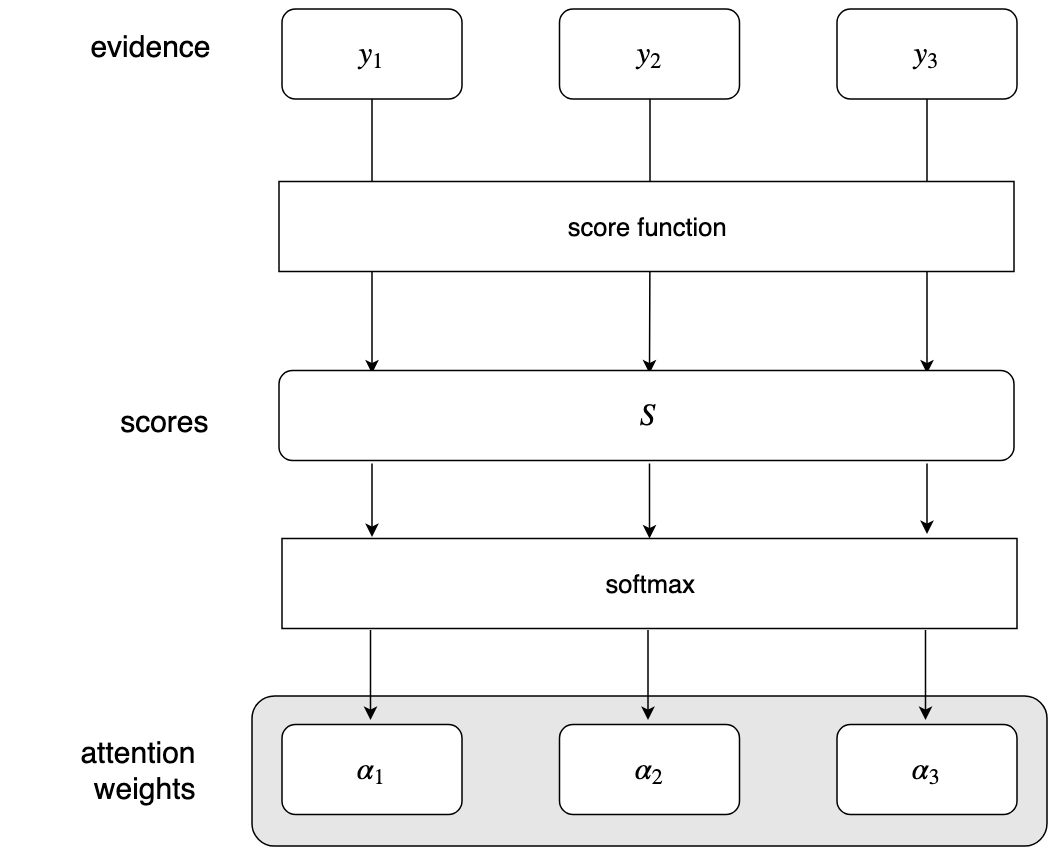}
\caption{Computing the attention weights. Given some evidence $y_i$ we apply our score function and produce the score set $S$. We then apply softmax and receive a distribution $\alpha_1, \alpha_2, \alpha_3$ which we call attention weights.}
\label{fig:attentionweights}
\end{figure}

Let us walk through an example. Consider a sequence $X = [x_1, x_2, x_3]$, with the corresponding evidence vector $Y = [y_1, y_2, y_3]$. Suppose that a score function $s$ is given. Let us proceed to calculate attention.
First, we must compute our attention weights. Our score function takes our evidence and produces a set of scores $S$, see Figure~\ref{fig:attentionweights}. We take a softmax of this set and get three attention weights: $\alpha_1, \alpha_2, \alpha_3$.

\begin{figure}[htbp]
\centering
\includegraphics[width=0.7\textwidth]{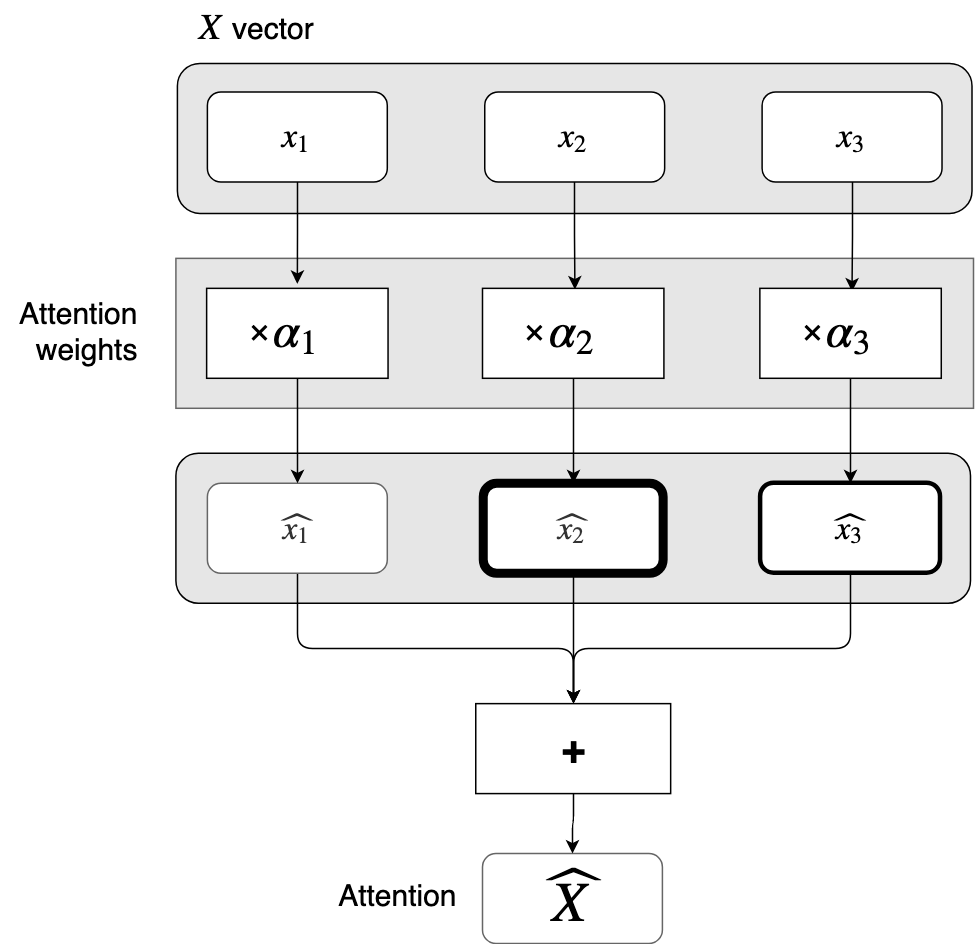}
\caption{Computing attention. Starting from the top we have: the original vector of vectors $X = [x_1, x_2, x_3 ]$, the multiplication by attention weights $a_i$'s, the weighted $\hat{x_i}$'s and following a sum the final attention value $\hat{X}$. As an example, we make the border thickness differ for each $\hat{x_i}$ to empathize that a corresponding $x_i$ got more attention (thick border) or less attention (thin border).}
\label{fig:attending}
\end{figure}

Now that we have our attention weights, we can proceed to compute attention, see Figure~\ref{fig:attending}. We start by multiplying our original elements $x_i$ by the attention weights. This results in vectors $\hat{x}_1, \hat{x}_2, \hat{x}_3$, which are the weighted versions of vectors $x_1, x_2, x_3$. To emphasize this we make the borders on some of the boxes thicker, representing a bigger weight. Intuitively this is what we refer to as ``how much attention'' is paid to a particular value $x_i$. Finally, we sum all the $\hat{x_i}$ vectors and get our attention value $\hat{X}$.

\section{Related Work}
\label{sec:relatedwork}
The authors in \cite{schnober2016comparing} compare three different variants of encoder-decoder models: two attention-based models and a character sequence to sequence model. They use data from Twitter Typo Curpose and Combilex data set which contains English graphemes to phonetic representation mappings. They compare these models against specialized monotone seq2seq models, and find them to be comparable in terms of performance. However, they did not look at the transformer or CNN architectures, nor did they use the same sentence input-output scheme that we propose here. In \cite{chaudhari2019attentive} authors propose taxonomy of attention where they classify different models based on number of sequences, number of abstraction levels, number of positions and number of representations. They also point out the distinct network architectures that use attention. These include: encoder-decoder, memory networks, networks with RNNs. In \cite{galassi2019attention} authors group attention models based on ``tasks addressed'' classification. These include things like document classification, machine translation, morphology, information extraction, syntax etc. A ``unified attention model'' is introduced which is similar to Transformer model. The authors propose a taxonomy of attention models based on input representation, compatibility functions, distribution functions, multiplicity.

Finally, we note that none of the related work we saw ever explicitly mathematically defined attention, like we do here, without any reference to the RNNs or sequence to sequence models.

\section{Method}
\label{sec:method}
In this section we describe the various models that we tested. We start with simple Sutskever model that does not use attention. Then we talk about Bahdanau model which was one of the first to introduce the notion of attention into sequence to sequence learning. Following that we talk about convoluational attention model, and finish by describing the current state of the art attention approach --- the transformer model.

\subsection{Sutskever model without attention}
In \cite{sutskever2014sequence} Sutskever et. al. use the encoder decoder architecture, and propose the use of some fixed dimensional representation of the input sequence to initialize the decoder. Consider the input sequence $x_1, \ldots x_T$ and a corresponding output sequence $y_1, \ldots, y_{T'}$. Authors first obtain representation $v$ of the input sequence (by simply taking the last hidden state of the LSTM encoder) and use that to predict the next item in the output sequence. In other words, having $v$ is fully sufficient to predict the next $y$:

\begin{equation}
p(y_t | x_1, \ldots, x_T) = p(y_t | v, y_1, \ldots, y_{t-1} )   
\label{eq:sutskever}
\end{equation}

\begin{figure}[htbp]
\centering
\includegraphics[width=0.6\textwidth]{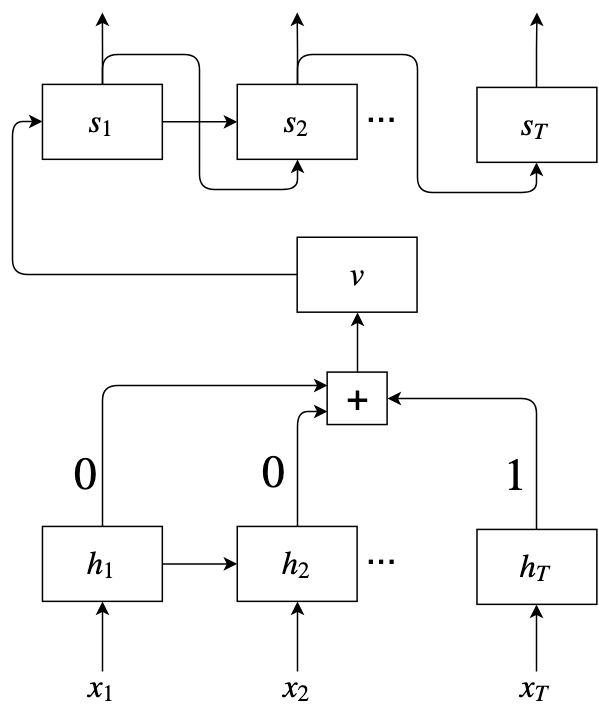}
\caption{Sutskever sequence to sequence model re-formulated as an attention model. Here the $x_i$'s are the original words in the sequence, the bottom part of the diagram represents the encoder component and the top part represents the decoder component. The $h_i$'s are the hidden states in the encoder, the $s_i$'s are the hidden states of the decoder, the numbers $0$ and $1$ next to the arrows going into the $+$ node represent the attention weights. The $v$ is the resulting \emph{context vector} ($\hat{X}$ in Equation~\ref{eq:attention}) or fixed dimensional representation, which is then fed into the decoder as the initial hidden state.}
\label{fig:seq2seq}
\end{figure}

In our Equation~\ref{eq:attention} this model reduces to a trivial attention case of having the scoring function assign all of the attention to the final component of the $X$ vector. As can be seen in Figure~\ref{fig:seq2seq} there are  hidden states $h_i$ which we can use to define:

\[
    X = [h_1, \ldots, h_T]
\]

So that $X$ is the vector of all hidden states of the encoder LSTM. We then attend to the last component, resulting in:

\[
  \hat{X} = X\cdot [0, \ldots, 1]^T = h_T = v
\]

In other words, the fixed representation $v$ used in Sutskever model is exactly the attention as we define it in Equation~\ref{eq:attention}. 

\subsection{Bahdanau model with attention}
\label{sec:bahatt}
 In \cite{bahdanau2014neural} Bahdanau proposes using attention over the encoder's hidden states.
 
Attention based models \cite{bahdanau2014neural} learn to align the relevant information extracted from the input sequence with the relevant information needed to produce a given word in the output sequence.

Whereas in the Sutskever model above each new item in the sequences was generated using the same fixed dimensional representation $v$ (see above), here the authors make this representation vary. Instead of $v$ they use $c_i$ which stands for context vector. The next word to be generated then depends on this custom, per step, context vector:

\[
    p(y_t | x_1, \ldots, x_T) = p(y_t | c_t, y_1, \ldots, y_{t-1})
\]

As we can see, the difference from Equation~\ref{eq:sutskever} is very small, but the interesting part is computing the context vector $c_t$. 

Let us use the convenient notation $t$ to refer to a timestep or index in the output sequence, and $t'$ to refer to a timestep or index in the input sequence:

\[
    c_t = \sum_{t'=1}^{T_x} \alpha_{t, t'} h_{t'}
\]

where the $h_{t'}$ are the ``annotations'' or the hidden states of the encoder, and $\alpha$'s are attention weights, defined as a softmax over some \emph{scores} $e$ (which we describe below):

\[
    \alpha_{t, t'} = \frac{\exp(e_{t, t'})}{\sum_{k=1}^{T_x} \exp(e_{t, k})}
\]

What are these scores? They are simply some function of the previous hidden decoder state $s_{t-1}$ and the targeted hidden encoder state:

\[
    e_{t, t'} = f(s_{t-1}, h_{t'} | \theta)
\]

This $f$ is parameterized by some $\theta$ which enables us to learn it. 

\begin{figure}[htbp]
\centering
\includegraphics[width=0.7\textwidth]{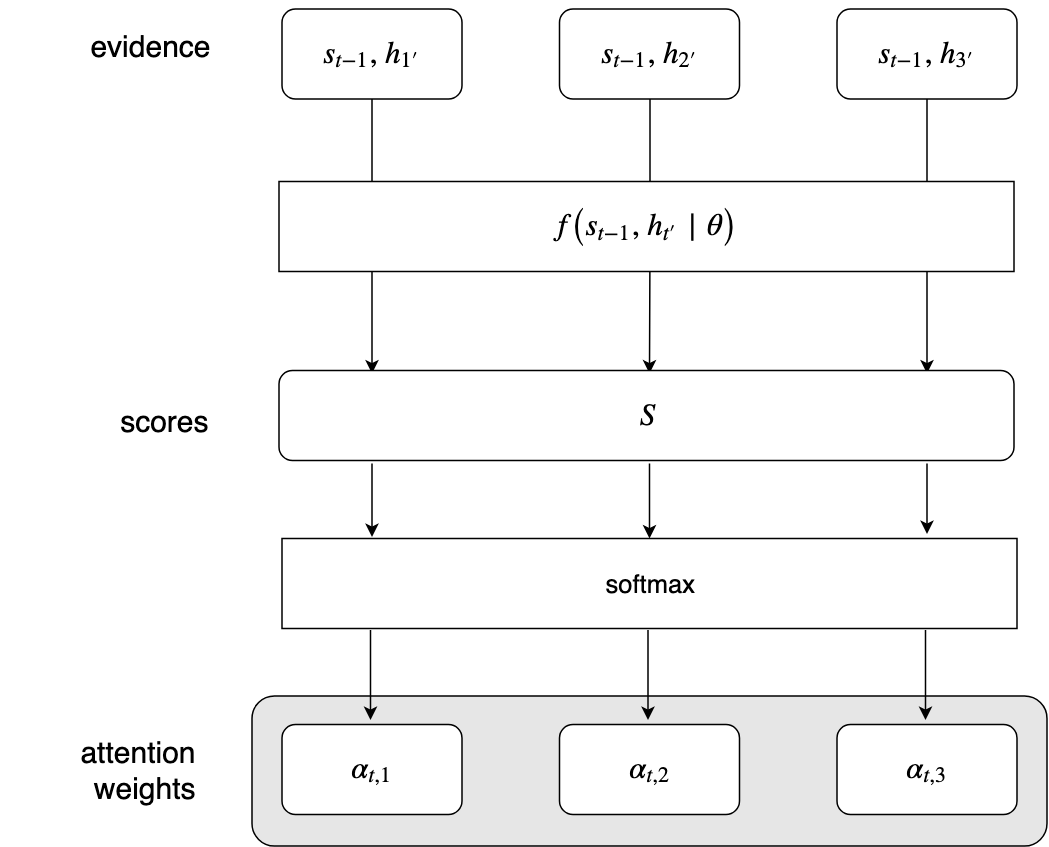}
\caption{Computing attention weights in Bahdanau model for step $t$ using the previous decoder state $s_{t-1}$ and all the encoder's hidden states $h_{t'}$. Notice that the only thing that varies in time here is the decoder's state $s_{t-1}$, which in turn produces new set of scores $S$ and thus a new set of attention weights.}
\label{fig:bahdanau_attention_weights}
\end{figure}

\begin{figure}[htbp]
\centering
\includegraphics[width=0.7\textwidth]{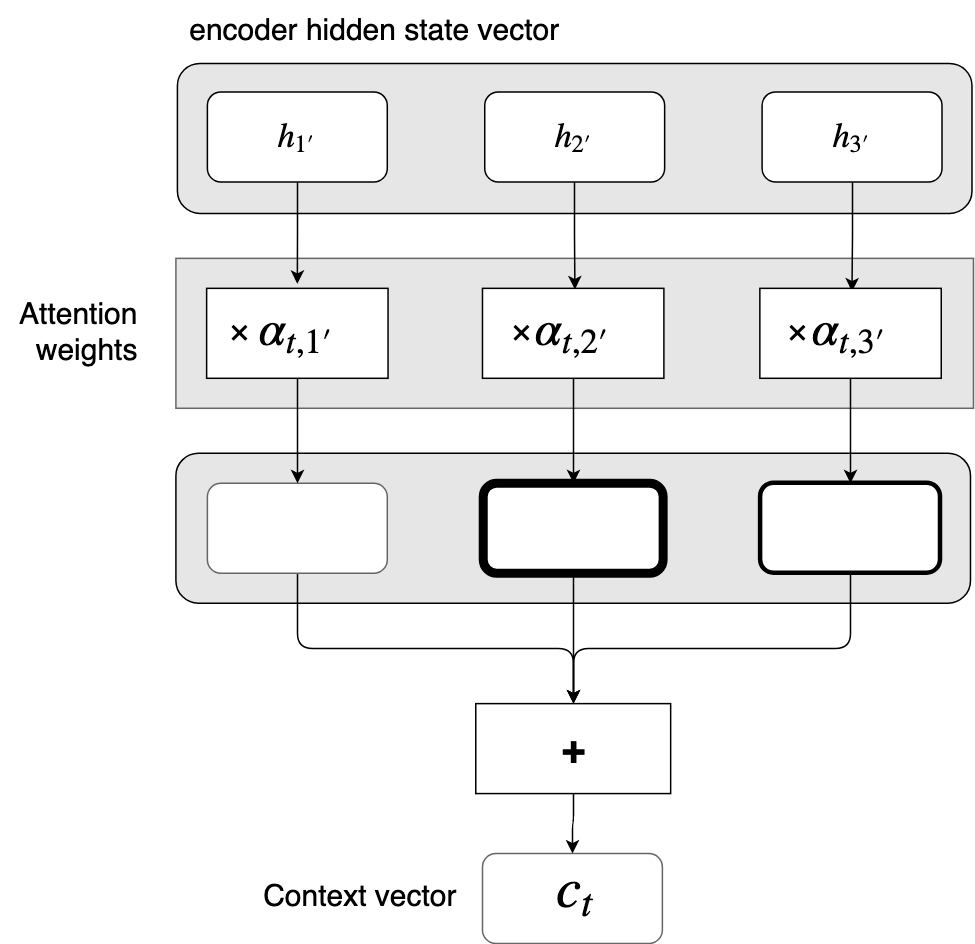}
\caption{Computing attention in Bahdanau model for step $t$ using computed weights. Here, for each $t$, we always use the same hidden states of the encoder $h_{t'}$ but use our previously computed attention weights $\alpha_{t,i'}$ for $i=1, 2, 3$. Doing the dot product between these vectors results in the context vector $c_t$ (which we know as \emph{attention}) for step $t$.}
\label{fig:bahdanau_attention}
\end{figure}

We can easily update the attention figures with the Bahdanau attention. The results in first computing the attention scores in Figure~\ref{fig:bahdanau_attention_weights} and then using them to compute a context vector for a particular step in Figure~\ref{fig:bahdanau_attention}.

\subsection{Convolutional Sequence to Sequence}
\label{sec:cnnmodel}
\begin{figure}[htbp]
\centering
\includegraphics[width=0.9\textwidth]{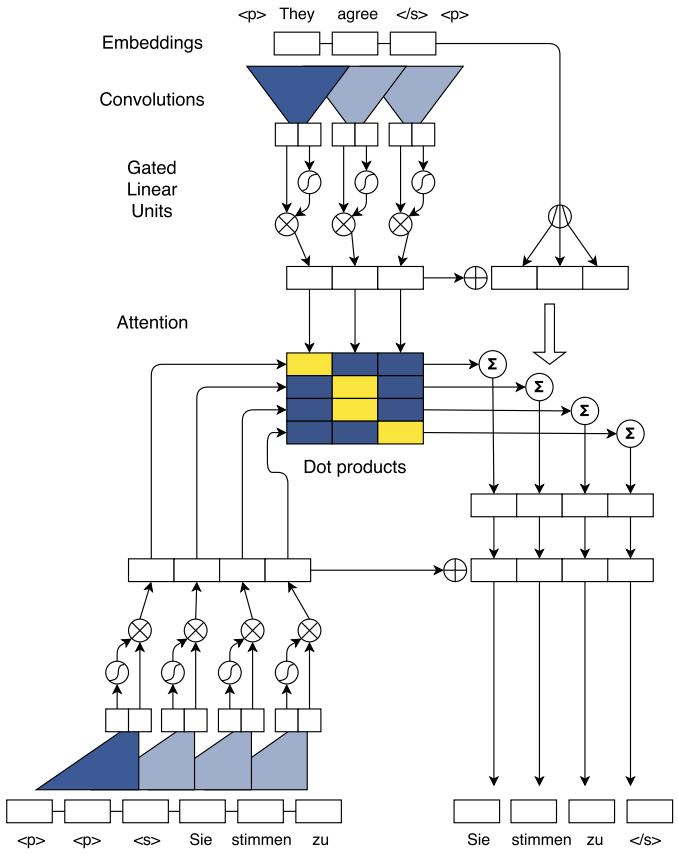}
\caption{Convolutional attention model. The input sentence is at the top and that output sentence as at the bottom right. The center component $4\times 3$ blocks in size are the attention weights. The $\Sigma$ components are the dot products between attention weights and the values coming from the top (indicated via a downward arrow). The final attention is the four block layer just below this operation.}
\label{fig:cnn_model}
\end{figure}

The method described in Section \ref{sec:bahatt} significantly improved the results of sequence generation compared to the previous works. Using recurrent networks has its drawbacks though. RNNs are very slow to compute since they work iteratively which means that in order to compute the next output you must first compute all the previous outputs. Because of this many researchers focused creating a network that can efficiently use the whole input sentence in less than $\mathcal{O}(N)$ steps. Gehring \etal in \cite{gehring2017convolutional} proposed to replace the RNN module by  CNNs as seen in Figure \ref{fig:cnn_model}. 

An advantage over RNN is that CNN learns fixed length context representation and stacking multiple layers can create larger context representation. This gives CNN control over the maximum length of dependencies that it needs to consider. In RNN this length will depend on input sequence and can get very large. Not being dependent on previous time step, CNN can parallelize entire learning process and can be very fast to train in contrast to RNN where current step depends on the vector from previous step which results in sequential learning.

CNN allows a hierarchical learning process where in the lower layers the nearby input sequences interact and as we move down through the layers the interactions between distant input sequences are learned.

It can be shown that CNN's hierarchical learning process takes $\mathcal{O}(N/k)$ operations where $k$ is the number of layers and $N$ is the number of words in the input sequence, whereas  RNN would take $\mathcal{O}(N)$.

Inputs to CNN are fed through constant number of kernels and non-linearities, whereas RNN applies $N$ operations and non-linearities to the first word and only a single set of operations to the last word. Fixed number of non-linearities in CNN expedites the training process.

The proposed method applies separate attentions to each decoder layer and demonstrates that each attention adds very little overhead. Specifically, the model introduces a separate attention at each decoding step using current hidden state $h_i^l$ of the decoder and previous embedding $g_i$. 

\[
d_i^l = W_d^lh_i^l + b_d^l + g_i
\]

For decoder layer $l$ the attention weight $\alpha_{ij}^l$ of state $i$ and source element $j$ is calculated as a dot product between the decoder state summary $d_i^l$ and each out of the last decoder step $z_j^u$. 

\[
    \alpha_{ij}^l = \frac{\exp(d_i^l \cdot z_j^u)}{\sum_{k=1}^{m} \exp(d_i^l \cdot z_k^u)}
\]

The attention, as per our Equation~\ref{eq:attention}, is the conditional input $c_i^l$ to the current decoder layer, and is a weighted sum of the encoder outputs as well as the input element embeddings $e_j$ (Figure \ref{fig:cnn_model}, center right):

\[
c_i^l = \sum_{j=1}^m \alpha_{ij}^l(z_j^u + e_j)
\]
 Once $c_i^l$ has been computed, it is simply added to the output of the
corresponding decoder layer $h_i^l$.

This can be seen as attention with multiple ``hops'' compared to single step attention described in Section \ref{sec:bahatt}.  In particular, the attention of the first layer determines a useful source context which is then fed to the second layer that takes this information into account when computing attention etc. This way allows for computation of attention across all elements of the sequence, compared to RNN where attention is computed sequentially for each step in the decoder.

\subsection{Transformer model}
Sequential nature of the RNNs is a problem in itself because it precludes parallelization. To solve this the authors in \cite{vaswani2017attention} use a transformer model. Transformer model parallelizes the attention step and allows the model to learn much faster.

Transformer is composed of encoder and a decoder. Both of these contain multiple layers. Each layer contains two components: attention mechanism and a feed-forward network. The attention mechanism in transformer, at first glance, looks rather different from what we have seen before. First we have the input which consists of queries $Q$, keys $K$ and values $V$. These in turn get multiplied together to yield the attention $A$:

\[
    A(Q, K, V) = \textit{softmax}\left(\frac{QK^T}{\sqrt{d_k}}\right) V
\]

\begin{figure}[htbp]
\centering
\includegraphics[width=0.8\textwidth]{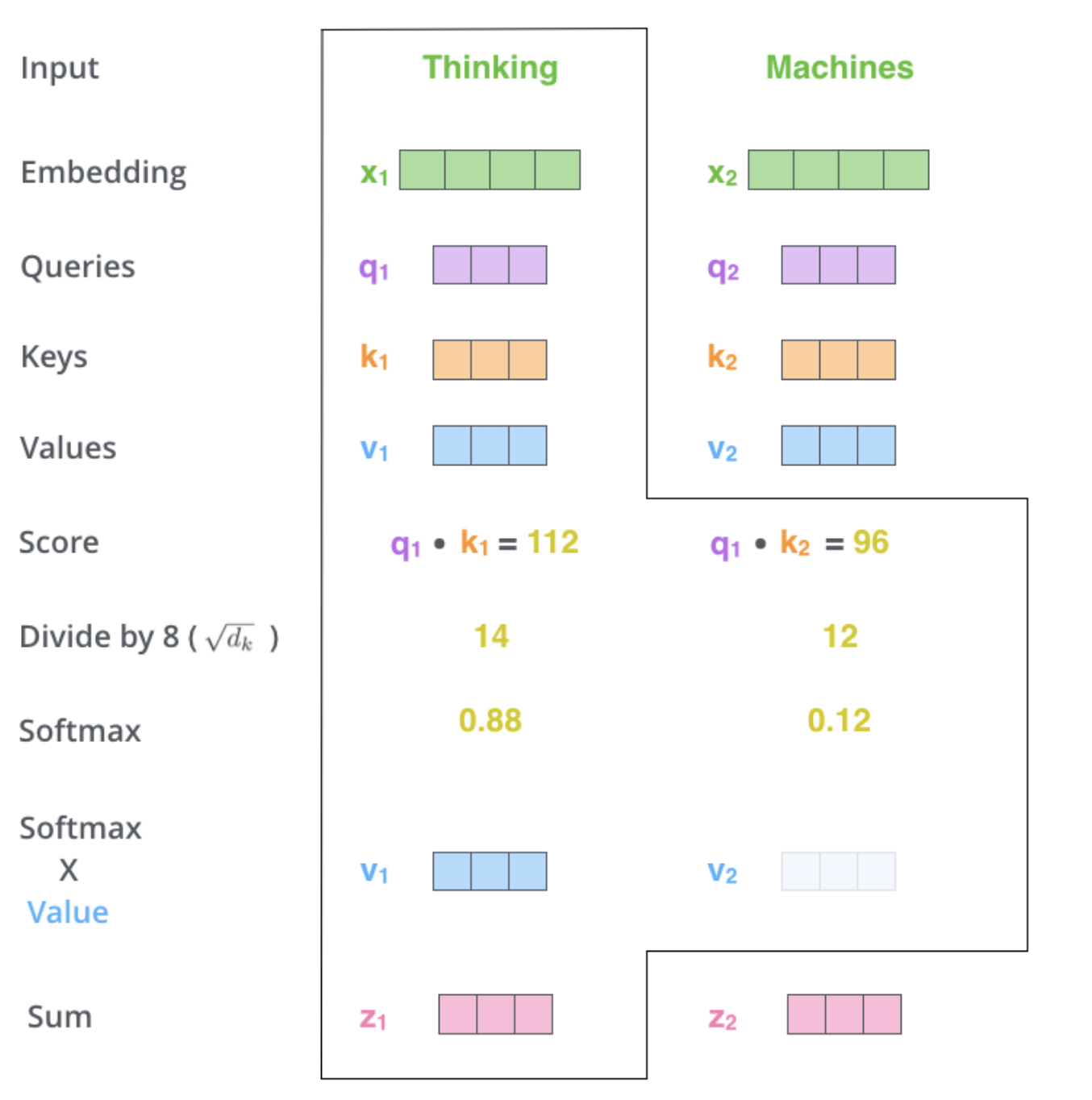}
\caption{Example of how transformer Q, K, and V values are computed from original X. Image taken from  \cite{jalamar2018illustrated}.}
\label{fig:transformer_values}
\end{figure}

Here is how $Q, K, V$ values are obtained (see Figure~\ref{fig:transformer_values} for  an example):

\begin{gather*} 
Q = XW^Q \\ 
K = XW^K \\
V = XW^V
\end{gather*}

where $W^i$ matrices are simply some weights that are learned through backpropagation.

How can we view this in a form that agrees with Equation~\ref{eq:attention}? Let our set $X = V$, the set of values. Let our $Y=X$ with the score function being:

\[s(y_i; \theta) = s(x_i; \theta) = \frac{q_i\cdot k_i}{\sqrt{d_k}}; y_i \in Y, q_i \in Q, k_i \in K
\]
where $Q$ and $K$ are computed from $X$ as before and the parameter $\theta = \{W^Q, W^K\}$.

Then our scores are simply $S = \{s(x_i) | x_i \in X\}$.

We can get our final attention by taking the softmax and multiplying by our values:

\[
    \hat{X} = X\cdot \textit{softmax}(S) = V\cdot \textit{softmax}(S) 
\]

Note that since we have $X = Y$ this is \emph{self attention}.

On the decoder side, we use ``encoder-decoder attention'' instead, with the only difference being that we use encoder's keys $K$ and values $V$ during the score computation, while using $Q$ from the decoder output. This allows the decoder to always see all of the values in the encoded sequence.

For the ``self-attention'' part of the decoder, we don't want to allow the decoder to refer to any of the positions after the position of the word that it has currently produced. The simple way of implementing this is to set the scores for these positions to very low values, the authors set them to $-\inf$. This results in softmax always making them almost zero, thus eliminating their contribution to the attention output.

Finally, the paper further refined the self-attention layer by adding a mechanism called ``multiple-heads'' attention. Each of these heads is a linear transformation of the input representation. This is done so that different parts of the input representation could interact with different parts of the other representation to which it is compared in the vector space. This provides the model with the ability to capture various different aspects of the input and to improve its expressive ability. Essentially, the multi-head attention is just several attention layers stacked in parallel, with different linear transformations of the same input.

\section{Experiment}
\label{sec:experiment}
In this section we describe the dataset we used, the experiments that we ran and our simple evaluation methodology.

\subsection{Dataset}
Our dataset is a small portion of the \emph{One Billion Word Benchmark} \cite{chelba2013billion}. We randomly sampled 10,000 sentences to use as our training set and 1,000 sentences as our testing set, with a maximum length of 30 words. 

\subsection{Evaluation Methodology}
In our experiment we make the input sentence also be the target sentence. In other words, given a sentence $S$ all our models should output the same sentence $S$. This allows us to focus on comparison between models and not on worrying about inaccurate human labelling or evaluation of translation.

We use the BLEU score to compute the similarity between the target sentence and the one generated by the model. We make a design decision to ignore the end of sentence (EOS) tokens, because appending many EOS tokens results in higher BLEU score. In testing we found this works very well with no unexpected results.

We use PyTorch 1.0.1 deep learning framework for learning and testing our models.
\subsection{Parameters}

\subsubsection{Sutskever model}
\begin{figure}[htbp]
\centering
\includegraphics[width=0.5\textwidth]{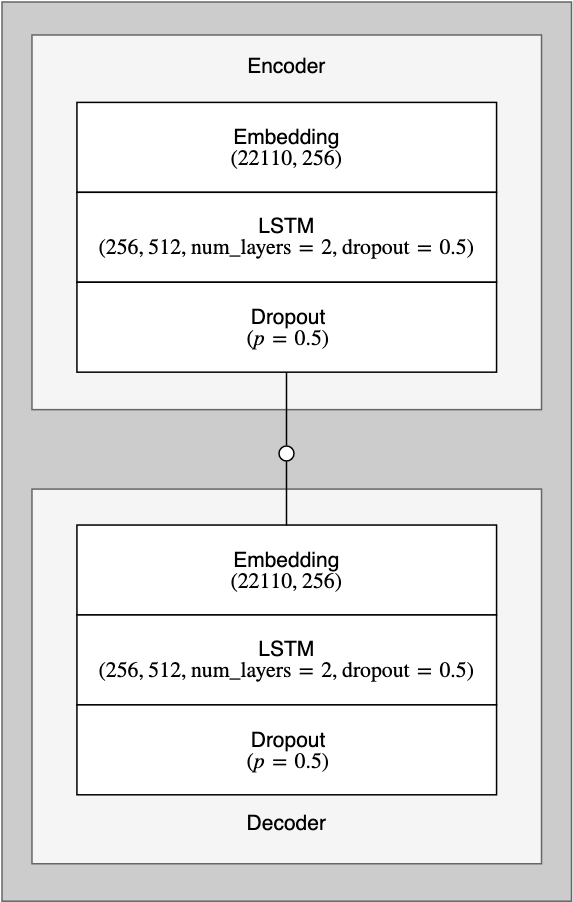}
\caption{The architecture for Sutskever model without attention. Embedding has a dimension of 256, both LSTM networks are the same and the final layer is a dropout layer. The first dimension here is a batch dimension, as is the convention in many deep learning frameworks.}
\label{fig:sutskever_architecture}
\end{figure}

For the simple case of no attention we use LSTM for both encoder and decoder parts of the model. See Figure~\ref{fig:sutskever_architecture}. We use embedding layers of dimension 256. The LSTMs are both with a hidden state depth of 512. We additionally set dropout to 0.5 during training on LSTMs. Finally, we use a dropout of 0.5 as the final layer in both encoder and the decoder.

We run the training indefinitely until we spot a trend of diminishing returns. In all other models that is about 50 epochs but for LSTM we choose to keep significantly more---700 epochs. We discuss the reasons in our results section.

\subsubsection{Bahdanau model}
\begin{figure}[htbp]
\centering
\includegraphics[width=0.5\textwidth]{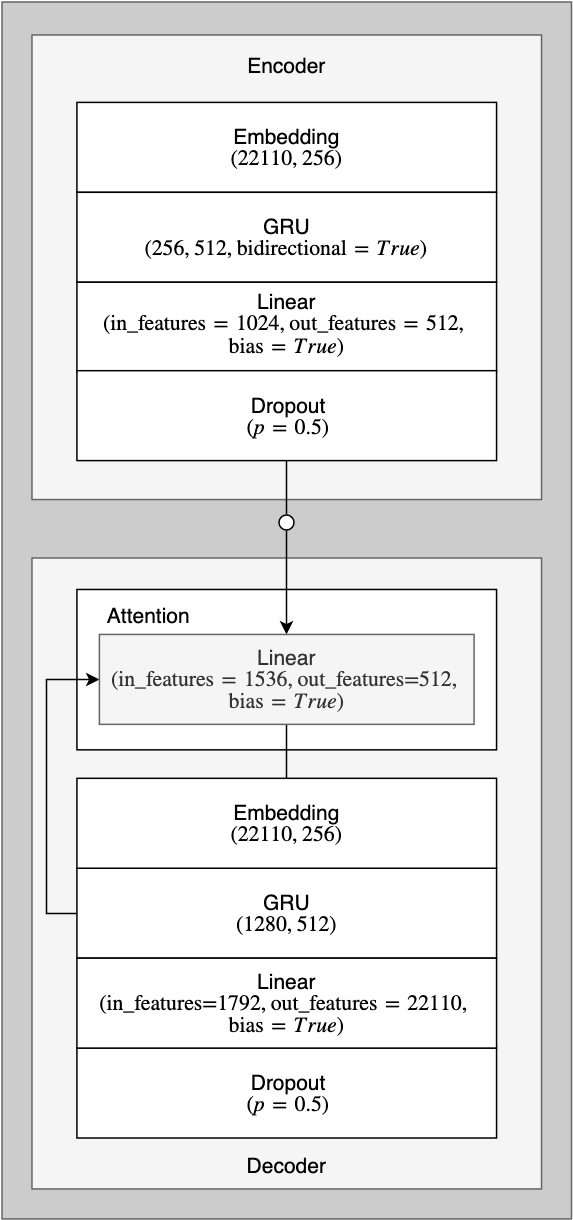}
\caption{The architecture for Bahdanau attention.}
\label{fig:bahdanau_gru_architecture}
\end{figure}

For the architecture of Bahdanau attention model we use a two component model with encoder/decoder. We chose to replace the LSTM recurrent network with the GRU \cite{DBLP:journals/corr/ChoMBB14} recurrent network in both components (See Figure~\ref{fig:bahdanau_gru_architecture}). According to \cite{DBLP:journals/corr/ChungGCB14}, GRU has comparable performance to LSTM but it uses less parameters and thus has faster running time. The encoder has an embedding layer of size 256. In the encoder we use a bidirectional GRU network with a hidden size of 512. We follow this layer with a fully connected layer that takes in all 1024 outputs (because GRU is bidirection we have 512*2=1024 outputs) from the GRU and outputs 512 features.

The network of the decoder is similar to the network of the encoder except that is has the new proposed attention layer and the GRU component is not bi-directional. This attention layer consists of a fully connected layer of size 512. The input dimension to this layer is $(\text{encoder\_hidden\_dim} * 2) + \text{decoder\_hidden\_dim} = 1536$ because according to \ref{sec:bahatt} the scores $e$ are calculated by concatenating the current decoder hidden state to each of the encoder output representations. Finally, we use a dropout of 0.5 as the final layer in both encoder and the decoder.



\subsubsection{CNN model}
For the CNN model, both the encoder and the decoder have an embedding layer of size 256. Similarly, we chose to set the hidden dimension to 512. The positional embeddings are set to be 100 learnable vectors. CNN model has the drawback that it requires a fixed maximum sentence length. A linear layer is then used to convert the vectors from the embedding space to the hidden space. In addition, we chose 10 1-D CNN layers of kernel size 3 and padding of 1 for both the encoder and decoder. The number of input channels is 512, same as the hidden dimension and the output channels is two times the input channels. For both the encoder and decoder, after the CNN layers we used a second linear layer to convert the vectors from the hidden space back to embedding space as described in \cite{gehring2017convolutional}.

In the decoder network, we used an attention layer of input size 512 and output size 256. This layer is applied on top of every convoluted output of every CNN layer and the output of this attention layer is added to the previous embeddings produced by the encoder (as described in Section \ref{sec:cnnmodel}. Once $c_i^l$ is computed, another linear layer is used to bring back the context vector to the hidden size space. Finally, we use a dropout of 0.25 as the final layer in both encoder and the decoder.

\subsubsection{Transformer model}
For the transformer model, we chose the hidden dimension to be 512 for consistency with the previous models. In addition, we chose to use 6 transformer layers for both the encoder and decoder. We also set the number of head on the ``multiple-heads'' attention to be 8. The inner-layer dimensionality of the position-wide feed-forward layer was set to 2048. Finally, we use a dropout of 0.1 as the final layer in both encoder and the decoder.

All the aforementioned models are optimized using Adam optimizer \cite{Kingma2015AdamAM} with default settings except for the Transformer model which is trained using the optimization method described in Section 5.3 in \cite{vaswani2017attention}.

\subsection{Results}

\begin{figure}[htbp]
\centering
\includegraphics[width=\textwidth]{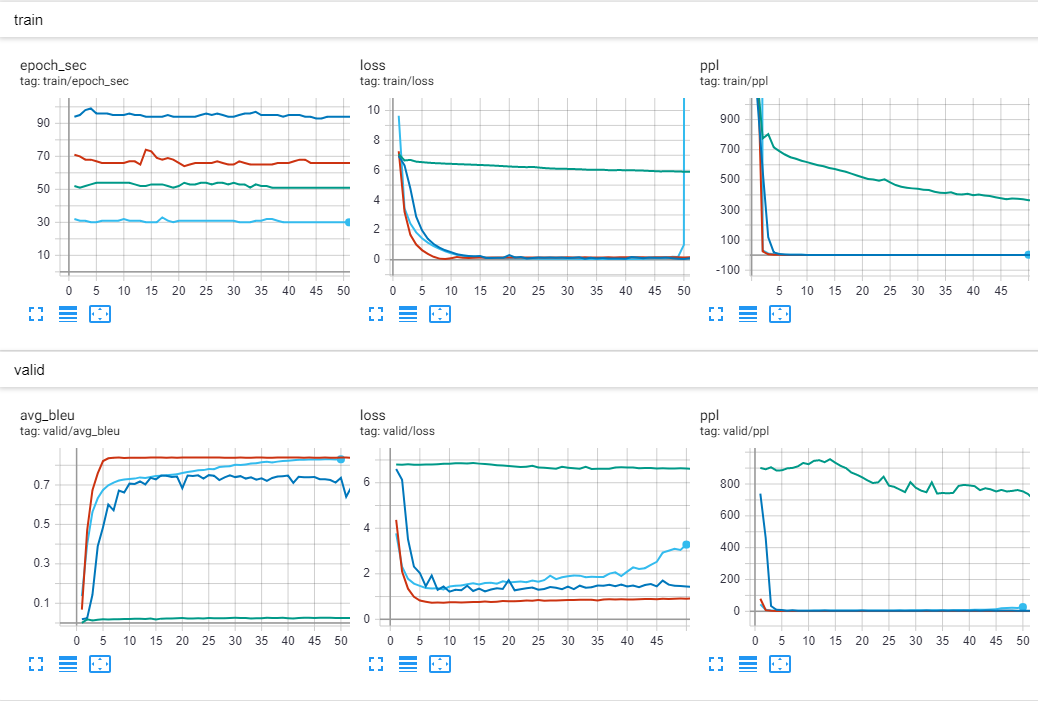}
\caption{All the models overlaid on the same graph. Looking at the bottom-left graph, the models are: \textcolor{red}{Transformer} (red), \textcolor{cyan}{CNN} with attention (cyan), \textcolor{blue}{GRU} with attention (blue) and \textcolor{green}{LSTM} without attention (green). The BLEU score of CNN approaches that of transformer, but the diminishing training gradient of the CNN caused a loss explosion (middle top graph) and further training failed.
Top panel: \emph{epoch\_sec} - training time per epoch, \emph{loss} -- training loss, \emph{ppl} -- training perplexity. Bottom panel: \emph{avg\_bleu} -- average epoch BLEU score, \emph{loss} -- testing loss, \emph{ppl} -- testing perplexity.}
\label{fig:overlay}
\end{figure}

Of all the models tested we find that Transformer is the most accurate and the fastest to converge. Refer to Figure~\ref{fig:overlay} which shows all the metrics overlaid on top of each other. The figure is divided into two panels: training metrics in the top panel, and testing metrics (valid) in the bottom panel.

The \emph{train} top panel contains the graphs for the following:

\begin{enumerate}
    \item \emph{epoch\_sec} - How many seconds did it take to train on one epoch?
    \item \emph{loss} - The loss computed by each model.
    \item \emph{ppl} - Perplexity measure.
\end{enumerate}

As we can see the CNN takes the least time to process one epoch worth of data, while the GRU takes the longest. The LSTM training loss, which is our model without attention, practically does not decrease. Note the sharp rise in the CNN loss (right part of the middle top graph). Training CNN past this point is not possible due to vanishing gradients. Similar problem---vanishing gradient and the loss explosion---is present in the GRU training, but occurs at a later point, at epoch 158. GRU does not crash, and the training proceeds, slowly recovering (we do not show this graph because of the space constraints).

The \emph{valid} bottom panel represents the testing metrics:

\begin{enumerate}
    \item \emph{avg\_bleu} - Average BLEU score computed over the epoch.
    \item \emph{loss} - Training loss computed by each model.
    \item \emph{ppl} - Perplexity measure.
\end{enumerate}

\begin{figure}[htbp]
\centering
\includegraphics[width=\textwidth]{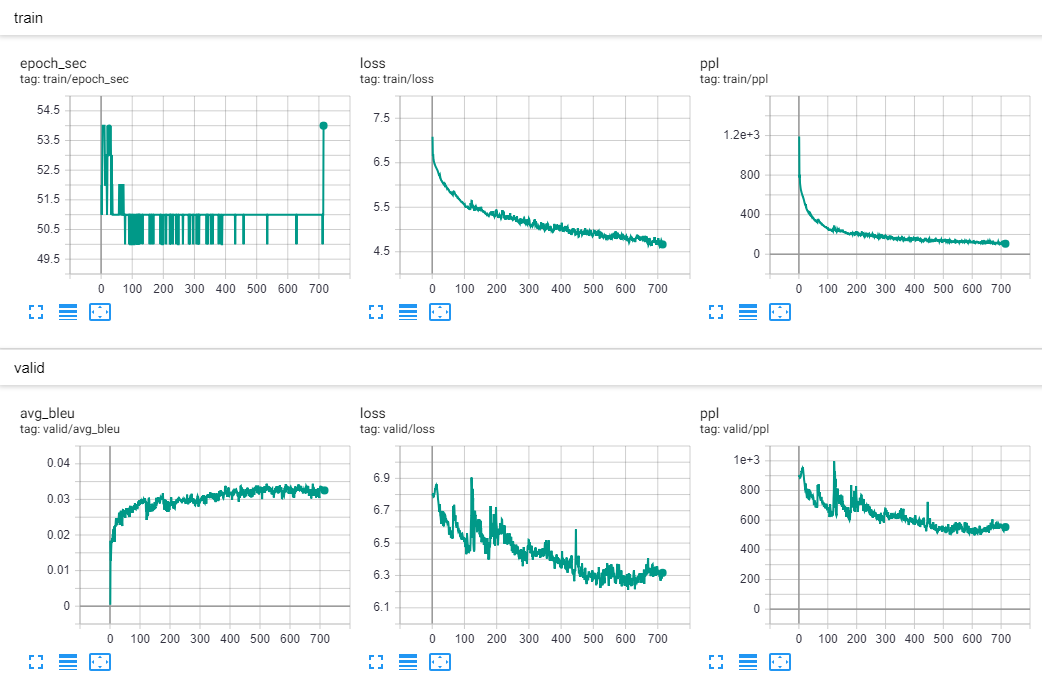}
\caption{Sequence to sequence model with LSTM. We can see that sequence to sequence model using a recurrent network takes over 500 epochs to stabilize. Even then, the improvment is negligible, barely reaching from 0.03 to 0.035. Top panel: \emph{epoch\_sec} - training time per epoch, \emph{loss} -- training loss, \emph{ppl} -- training perplexity. Bottom panel: \emph{avg\_bleu} -- average epoch BLEU score, \emph{loss} -- testing loss, \emph{ppl} -- testing perplexity.}
\label{fig:lstm}
\end{figure}

Transformer converges to its optimal value of BLEU score very quickly, just after about six epochs. The CNN does pretty good and even approaches the BLEU score of the Transformer, however it crashes during training when this happends due to the vanishing gradient issue we mentioned above. Next best performing model is the GRU with attention, which does about 0.1 BLEU points below that of a transformer, and always fluctuates in performance. It takes longer to reach its optimal value, about 15 epochs. However, after epoch 158, the GRU model experiences a significant decrease in BLEU score, going down to 0.02 after the vanishing gradient and loss explosion presents itself. 
Finally, the plain LSTM model performs the worst and never even breaks the 0.1 BLEU score. This is also the case if we let it run for much longer than other models, say 700 epochs (see Figure~\ref{fig:lstm}). On the other hand, it never experiences the drastic crashes of GRU or CNN models.

We record the best BLEU scores achieved by each model in Table~\ref{table:results}. Note that while CNN achieves a similar BLEU score to Transformer, it does so much slower, taking over 50 epochs. Transformer is the clear winner, while taking not much longer to run than other models.

\begin{table}[htbp]
\centering
\caption{Best results for each model. BLEU score, seconds per epoch and number of epochs it took to converge to the best BLEU score.}
\begin{tabular}{@{}lllll@{}}
\toprule
Model       & BLEU            & Sec/epoch   & Epochs to converge & Num. of parameters \\ \midrule
LSTM        & 0.03            & 50          & \textgreater 500   & 30,019,166 \\
GRU         & 0.74            & 93          & 16                 & 57,397,854 \\
CNN         & 0.8302          & \textbf{30} & 50                 & 49,320,286 \\
Transformer & \textbf{0.8392} & 66          & \textbf{6}         & 79,127,134 \\ \bottomrule
\end{tabular}
\label{table:results}
\end{table}

\section{Discussion}
It seems that the primary difference between models like Transformer, attention-based  GRU or CNN and \emph{attention-less} LSTM is the degree to which the attention, or more specifically the scoring function, is learned. In the most basic case of LSTM where there is no attention, we observe the worst performance. This coincides with hard-coding the scoring function to be some constant. So it is that the most rudimentary version of attention, one where it is not learned---performs the worst. Next we have GRU and CNN with attention, which perform significantly better than LSTM. The difference being here that in both models the scoring functions have some parameters that are learned. This seems to allow the model to explore a much larger space of solutions and leads to better outcomes. The last restriction that we can remove is the sequential nature of the scoring function. Transformer does exactly that: the scoring function is not dependent on the previously computed state, and the \emph{temporal} nature of the sequence is now learned from the data itself (by encoding positional information into the data). The fact that transformer performs significantly better than previous approaches validates our hypothesis that the more the model is able to learn the scoring function, and thus the better it can make use of attention, the more effective the attention mechanism becomes.

In fact, attention mechanism seems to be a fundamental property of the systems that learn, much like backpropagation. It seems to us that one simply has to look for these two properties:

\begin{enumerate}
    \item Selecting some largest element from the set
    \item Evidence not necessarily part of the set we are selecting from
\end{enumerate}

and upon funding them implicitly encoded into the architecture of the model, letting them vary and be learned. This way we can \emph{add} a stronger attention mechanism to any model that already has some weak form of it.

In addition, using non-recurrent attention architectures like the CNN and the transformer not only improves the performance but also significantly reduces the running time for a single pass through the model, if we take into consideration the number of parameters each model has to learn (Table \ref{table:results}).

\section{Future work}
One possible direction in the future is to validate the benefit of attention mechanism in a more complex sequential or temporal domains like translation or question answering. Additionally, one could experiment with different modalities like computer vision, audio synthesis, recommendation systems, video generation and  stock market data. It may be that attention is equally applicable in these different domains or that it is not well suited to some.

It is also worth exploring the case of hard attention. This is when we replace the operation of softmax with argmax in the mathematical formulation of attention. Since we cannot compute the derivative of argmax due to discrete nature of this operation, it would be interesting to explore other methods of learning, such as reinforcement learning, for our objective function. This may result in better performance and speed.

Since CNN and transformers are both newly proposed methods and they work in a non-recurrent fashion, the position encoding that is given at the beginning is crucial in order to incorporate the ordering of the input sequence. Currently, all the proposed approaches use either a fixed representation of the position or a learned one. We think this positional representation can be improved by adding an additional attention layer acting on positional representation. This can be added before incorporating the positional representation with the input representation. The intuition behind this is that right from the beginning we give the model the power of generating more general positional representations according to the learned attention.

\section{Conclusion}
We found that transformer performs the best in the sequence to sequence mapping tasks. By making use of attention it is able to capture long range dependencies and consistently decrease the loss without showing any instability that other models show. Unlike transformer, the CNN and GRU attention models take much longer to train and produce worse results. Additionally, these models are unstable, meaning that over-training them causes unanticipated results. The LSTM model without attention shows practically no learning and should not be used for any practical applications.

In our opinion using a simple task, like copying a sentence, allowed us to isolate the effect that attention had on the performance of sequence to sequence mapping models. It is clear that attention is very effective in this regard. We have also hypothesized that even models \emph{without} attention can still be viewed as attention-based models but with predefined or non-learnable scoring functions.

Attention is a mechanism that exists outside machine learning and natural language processing. It is simply a mathematical operation of selecting the largest element from a set based on some evidence. Given such a general definition it is surprising how useful it turned out to be in many machine learning tasks. Translation is the most obvious application area but we think it might apply to many other domains for which attention has not been considered yet. However, by carefully looking for key ingredients of attention mechanisms in existing models we will see many more applications.

\bibliographystyle{unsrt}  
\bibliography{report}

\begin{thebibliography}{10}

\bibitem{graves2013generating}
Alex Graves.
\newblock Generating sequences with recurrent neural networks, 2013.

\bibitem{sutskever2014sequence}
Ilya Sutskever, Oriol Vinyals, and Quoc~V Le.
\newblock Sequence to sequence learning with neural networks.
\newblock In {\em Advances in neural information processing systems}, pages
  3104--3112, 2014.

\bibitem{vaswani2017attention}
Ashish Vaswani, Noam Shazeer, Niki Parmar, Jakob Uszkoreit, Llion Jones,
  Aidan~N. Gomez, Lukasz Kaiser, and Illia Polosukhin.
\newblock Attention is all you need, 2017.

\bibitem{bahdanau2014neural}
Dzmitry Bahdanau, Kyunghyun Cho, and Yoshua Bengio.
\newblock Neural machine translation by jointly learning to align and
  translate.
\newblock {\em arXiv preprint arXiv:1409.0473}, 2014.

\bibitem{gehring2017convolutional}
Jonas Gehring, Michael Auli, David Grangier, Denis Yarats, and Yann~N. Dauphin.
\newblock Convolutional sequence to sequence learning, 2017.

\bibitem{parikh2016decomposable}
Ankur~P. Parikh, Oscar Täckström, Dipanjan Das, and Jakob Uszkoreit.
\newblock A decomposable attention model for natural language inference, 2016.

\bibitem{schnober2016comparing}
Carsten Schnober, Steffen Eger, Erik-Lân~Do Dinh, and Iryna Gurevych.
\newblock Still not there? comparing traditional sequence-to-sequence models to
  encoder-decoder neural networks on monotone string translation tasks, 2016.

\bibitem{chaudhari2019attentive}
Sneha Chaudhari, Gungor Polatkan, Rohan Ramanath, and Varun Mithal.
\newblock An attentive survey of attention models.
\newblock {\em arXiv preprint arXiv:1904.02874}, 2019.

\bibitem{galassi2019attention}
Andrea Galassi, Marco Lippi, and Paolo Torroni.
\newblock Attention, please! a critical review of neural attention models in
  natural language processing.
\newblock {\em arXiv preprint arXiv:1902.02181}, 2019.

\bibitem{jalamar2018illustrated}
Jay Alammar.
\newblock The illustrated transformer.
\newblock \url{http://jalammar.github.io/illustrated-transformer/}, June 2018.

\bibitem{chelba2013billion}
Ciprian Chelba, Tomas Mikolov, Mike Schuster, Qi~Ge, Thorsten Brants, Phillipp
  Koehn, and Tony Robinson.
\newblock One billion word benchmark for measuring progress in statistical
  language modeling, 2013.

\bibitem{DBLP:journals/corr/ChoMBB14}
KyungHyun Cho, Bart van Merrienboer, Dzmitry Bahdanau, and Yoshua Bengio.
\newblock On the properties of neural machine translation: Encoder-decoder
  approaches.
\newblock {\em CoRR}, abs/1409.1259, 2014.

\bibitem{DBLP:journals/corr/ChungGCB14}
Junyoung Chung, {\c{C}}aglar G{\"{u}}l{\c{c}}ehre, KyungHyun Cho, and Yoshua
  Bengio.
\newblock Empirical evaluation of gated recurrent neural networks on sequence
  modeling.
\newblock {\em CoRR}, abs/1412.3555, 2014.

\bibitem{Kingma2015AdamAM}
Diederik~P. Kingma and Jimmy Ba.
\newblock Adam: A method for stochastic optimization.
\newblock {\em CoRR}, abs/1412.6980, 2015.

\end{thebibliography}

\end{document}